\documentclass[journal]{IEEEtran}
\usepackage{graphicx}
\usepackage{amssymb}
\usepackage{amsmath}
\usepackage{mathrsfs} % define this before the line numbering.
\usepackage{xcolor}
\usepackage{algorithm}
\usepackage{algorithmic}
\usepackage{makecell}
\usepackage{multirow}
\usepackage{booktabs}
\usepackage{flushend}
\usepackage{bm}
\usepackage[pagebackref=true,breaklinks=true,letterpaper=true,colorlinks, linkcolor=red,bookmarks=false]{hyperref}
\ifCLASSINFOpdf
\else
\fi
\begin{document}
\title{Multi-Granularity Canonical Appearance Pooling for Remote Sensing Scene Classification}

\author{Shidong~Wang,~\IEEEmembership{Student Member,~IEEE,}
	Yu~Guan, Ling~Shao,~\IEEEmembership{Senior Member,~IEEE}% <-this % stops a space
	\thanks{Manuscript received 26-Jul-2019; revised 15-Jan-2020 and 18-Feb-2020; accepted 18-Mar-2020. (\textit{Corresponding author: Ling Shao}).}
	\thanks{Shidong, Wang is with the School of Computing Sciences, University of East Anglia, Norwich, UK. (email:Shidong.Wang@uea.ac.uk).}% <-this % stops a space
	\thanks{Yu Guan is with the Open Lab, School of Computing, Newcastle University, Newcastle upon Tyne, UK. (email:Yu.Guan@ncl.ac.uk).}% <-this % stops a space
	\thanks{Ling Shao is with the Inception Institute of Artificial Intelligence, Abu Dhabi, United Arab Emirates (e-mail: ling.shao@ieee.org).}}


\markboth{IEEE TRANSACTIONS ON IMAGE PROCESSING}%Journal of \LaTeX\ IEEE TRANSACTIONS ON IMAGE PROCESSING,~Vol.~14, No.~8, August~2020
{Shell \MakeLowercase{\textit{et al.}}: Bare Demo of IEEEtran.cls for IEEE Journals}



\maketitle

\begin{abstract}
Recognising remote sensing scene images remains challenging due to large visual-semantic discrepancies. These mainly arise due to the lack of detailed annotations that can be employed to align pixel-level representations with high-level semantic labels. As the tagging process is labour-intensive and subjective, we hereby propose a novel Multi-Granularity Canonical Appearance Pooling (MG-CAP) to automatically capture the latent ontological structure of remote sensing datasets. We design a granular framework that allows progressively cropping the input image to learn multi-grained features. For each specific granularity, we discover the canonical appearance from a set of pre-defined transformations and learn the corresponding CNN features through a maxout-based Siamese style architecture. Then, we replace the standard CNN features with Gaussian covariance matrices and adopt the proper matrix normalisations for improving the discriminative power of features. Besides, we provide a stable solution for training the eigenvalue-decomposition function (EIG) in a GPU and demonstrate the corresponding back-propagation using matrix calculus. Extensive experiments have shown that our framework can achieve promising results in public remote sensing scene datasets.
\end{abstract}
\section{Introduction}
\label{intro}
Recently, the rapid development of remote sensing technology, especially in image acquisition equipment, has made it easy to obtain satellite images with high-spatial and high-spectral resolution. Gigabytes worth of remote sensing data is accumulated daily. This has led to a growing demand for intelligent algorithms that can analyse this valuable information. Successful algorithms are likely to be employed in a wide range of applications related to Earth observation. Examples include land use and land cover (LULC) determination, urban planning, vegetation mapping, natural hazard detection, environmental monitoring and Remote Sensing Scene Classification (RSSC)~\cite{datasets:xia2016AID, datasets:cheng2017Remote}.
\begin{figure}[]
	\centering
	\includegraphics[width=0.48\textwidth]{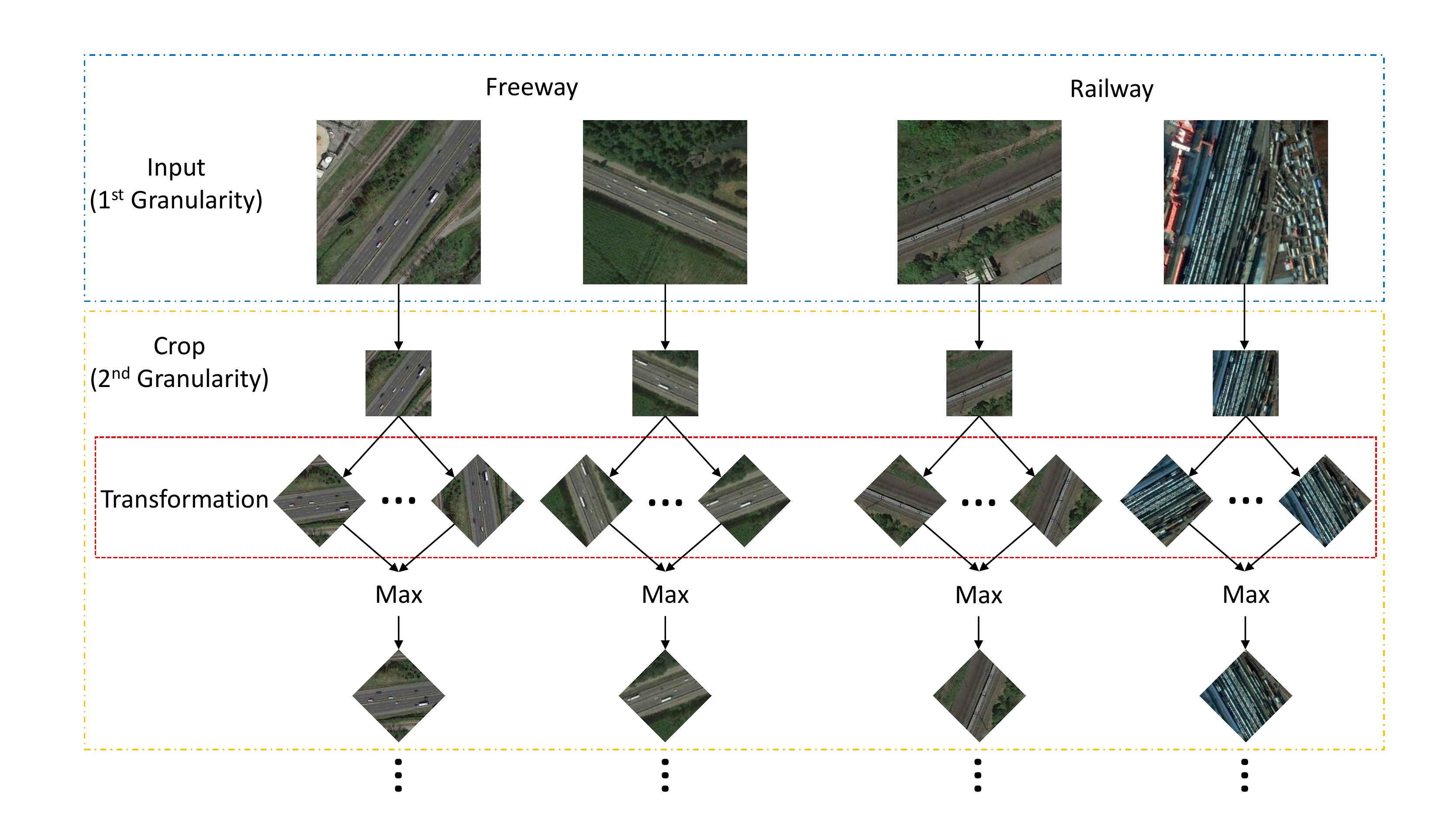}
	\caption{Example images selected from two different categories in the NWPU-RESISC45 dataset~\cite{datasets:cheng2017Remote}. To distinguish visually similar image, regions often need to be zoomed in to see the subtle differences. However, the differences are more significant and vivid if we transform the zoomed regions to their canonical appearances.}
	\label{intro_fig}
\end{figure}

Among the applications mentioned above, RSSC is one of the most active research fields in the remote sensing community. Specifically, RSSC is a process of classifying the remotely sensed images into discrete sets of LULC categories with semantic meanings. However, there are two main challenges in the RSSC that must be resolved. The most crucial problem is the large visual-semantic discrepancy that is caused by the lack of alignment between the visual features and semantic labels. Remote sensing scene images, covering a large geographic area with significant unstructured information, require different levels of annotation as supervision for classification. However, most existing remote sensing image datasets lack well-constructed ontological structures, with the result that high-level semantic meanings in category labels cannot be incorporated in the learned features. The second challenging problem is the variances that naturally appear in remote sensing scene datasets. Specifically, there are two major variances: intra-class diversity and inter-class similarity. As shown in Fig.~\ref{intro_fig}, the \textit{railway} image on the left is visually similar to \textit{freeway} images but is different from the \textit{railway} image on the right belonging to the same category. 

An intuitive way to address the above problems is to obtain detailed annotations. However, collecting well-annotated data is impractical because it requires massive amounts of manpower and is time-consuming and subjective. Such problems are more difficult to solve in remote sensing datasets since many categories have hierarchical ontologies. For example, in the NWPU-RESISC45 dataset~\cite{datasets:cheng2017Remote}, \textit{airplane} and \textit{runway} may belong to the same parent category \textit{airport}, similarly, \textit{railway} and \textit{railway\_station} may come from the category of \textit{railway} while \textit{bridge} pertains to \textit{freeway}. Moreover, \textit{airport}, \textit{railway} and \textit{freeway} are the branches of \textit{transportation}. In taxonomies, these relationships can be categorised into three levels based on the class inclusion and degree of specificity, which includes the superordinate-level, the basic-level and the subordinate-level~\cite{ungerer2013introduction,croft2004cognitive}. The further up in the taxonomy a category is located, the more general it is, and vice versa. Classifying the superordinate-level category \textit{transportation} or a basic-level category like  \textit{railway} and  \textit{airport} is relatively easy while more discriminative features are acquired for recognising subordinate-level classes, such as \textit{airplane} and \textit{runway}, \textit{railway} and \textit{railway\_station}, as well as \textit{bridge}. A similar hierarchical relationship can also be found among the categories in the AID dataset~\cite{datasets:xia2016AID}, such as \textit{dense residential}, \textit{medium residential} and \textit{sparse residential}.

Our framework builds on the assumption that there is a latent ontology between the basic-level and subordinate-level category labels in remote sensing scene datasets. As discussed earlier, the inclusion of latent hierarchical structures is a viable solution for decreasing the large visual-semantic discrepancy. However, manually designed ontologies are expensive to acquire and often suffer from subjectiveness. Therefore, we explore an alternative strategy to incorporate hierarchical information by learning granular feature representation. Notably, we expect the learned features to not only contain distinctive information from different granularities, but also are in line with the latent ontological structure of the datasets.

To achieve the target mentioned above, we propose a novel Multi-Granularity Canonical Appearance Pooling algorithm (MG-CAP) to learn granular feature representation for the RSSC tasks (see Fig~\ref{framework}). Different granularities are produced by progressively cropping the raw images multiple times. For each specialised grain level, multiple instances are generated based on a pre-defined set of transformations. Inspired by~\cite{chopra2005learning}, we apply a Siamese style architecture for feature extraction and learn the dependencies among different instances. Then, we collect second-order statistics of standard CNN features and convert them to Gaussian covariance matrices as the global representation. At the end of the Siamese architecture, we adopt the maximum operation to obtain a unique Gaussian covariance matrix with respect to the canonical appearance of the generated image instances. The learned feature is invariant to transformations and can mitigate the effects of large intra-class variations. The obtained Gaussian covariance matrices are Symmetric Positive Semi-definite (SPD matrices), which have been endowed with special geometric structures (\textit{e.g.,} pseudo-Riemannian manifold). We then implement a GPU-supported, non-linear EIG-decomposition function with suitable matrix normalisations to learn the pseudo-Riemannian manifold. Finally, we fuse the different grained features and feed the result into the classifier. Our contributions can be summarised as follows:
\begin{itemize}
	\item We derive a novel Multi-Granularity Canonical Appearance Pooling to incorporate the latent ontological structure of remote sensing scene datasets and then alleviate the impact of visual-semantic discrepancies.
	\item We progressively leverage Siamese style architectures to learn transformation-invariant features for solving the large intra-class variation problem.   
	\item We offer a stable GPU-supported EIG-decomposition function, which can conveniently exploit the Gaussian covariance geometry with different matrix normalisations.   
\end{itemize}     
\section{Related Work}
\label{rw}
\subsection{Handcrafted Feature-based Methods}
Handcrafted feature-based methods have been extensively explored to capture the low-level information of remote sensing scene images. Locally handcrafted features learn regional features by incorporating the transformation-invariant information, such as shape and the local structures.~\cite{yang2010bag} introduced a framework to improve the performance for RSSC tasks by employing the Scale Invariant Feature Transform (SIFT) feature. The SIFT feature is covariant to the re-scaling, translation and rotation of an image since it searches similar circular regions at multiple scales and positions. The HOG descriptor is another useful feature for capturing object shapes and edge structures in remote sensing images~\cite{shi2014ship,zhang2015generic,cheng2016object}. These low-level features are discriminative due to the collection of local structure information, but they usually require to be converted into intermediate descriptors to generate the entire representation.                                 

Globally handcrafted features can be directly fed into classifiers to generate the classification score. For example, the colour auto-correlogram (ACC) descriptor~\cite{huang1997image} aims to encode the spatial colour distribution by computing the probabilities of two pixels. Border-interior pixel classification (BIC)~\cite{stehling2002compact,dos2010evaluating,penatti2012comparative} computes colour histograms for both border pixels and interior pixels. The local colour histogram (LCH)~\cite{swain1991color} divides colour histogram into tiles and then calculates each histogram independently for the final concatenation. More recent works~\cite{li2010object,penatti2015deep} have achieved better performance than the conventional colour based methods, but are inadequate in conveying the spatial information and are sensitive to small illumination changes or quantisation errors. Besides, GIST feature-based methods~\cite{oliva2001modeling,li2011saliency,yin2015crater} represent the dominant spatial structures of a given image by computing a global holistic description.  

In addition, various texture descriptor based frameworks have been employed to depict the spatial arrangements and global statistics of aerial images. For example, in the early years, many approaches learned texture information by extracting Gabor features~\cite{jain1997object} or grey level co-occurrence matrix (GLCM) features~\cite{haralick1973textural}. Later, several works began to measure the spatial structure information of local image textures by employing local binary patterns (LBP)~\cite{ojala2002multiresolution,huang2016remote,ren2015learning}. Then, the local activity spectrum (LAS)~\cite{tao2000texture} achieved more reliable performance and performed more effectively in calculating the spatial activities in the vertical, horizontal, diagonal and anti-diagonal directions, respectively.     
\subsection{Mid-level Feature Learning-based Methods }
Handcrafted features can be fused to obtain a discriminative representation, but they rely heavily on prior human knowledge. The research trend in recognising remote sensing images has moved towards learning mid-level feature representations, often in the form of unsupervised learning-based methods. The popularity of unsupervised learning algorithms is primarily attributed to the reduction in labour-extensive image labels and effectiveness in removing the outliers of low-level features. In the following, we will elaborate on these algorithms from a linear perspective, while non-linear algorithms will be introduced in the section of deep neural network-based methods. 

Principal component analysis (PCA) is acknowledged as the simplest way to analyse the true eigenvector-based multivariates and the variants such as sparse PCA are commonly applied for the RSSC task~\cite{chaib2016informative}. K-means clustering is another unsupervised feature learning method, which aims to iteratively assign raw data to pre-defined centroids. The algorithm is a standard procedure for generating codebooks in bag-of-visual-words (BoVW) models~\cite{zhang2013high,zhao20142,bahmanyar2015comparative,zhao2016spectral,zhu2016bag}. Many researchers have considered sparsely encoding the high-dimensional features using structural primitives in a low-dimensional manifold~\cite{cheng2016survey}. For example,~\cite{bahmanyar2015comparative,cheriyadat2014unsupervised} generate a sparse vector using a group of learned functions from low-level features.~\cite{zheng2013automatic} imposes different constraints on features to jointly learn the sparse coding. Additionally, auto-encoder based methods~\cite{othman2016using,du2017stacked} adopt asymmetric neural networks to learn the compressed feature by minimising the reconstruction error between the input and the generated data. 
\subsection{Deep Neural Network based Methods}   
Deep neural network based methods have gained increasing popularity for remote sensing scene categorisation due to the efficiency of learning the expressive representation. Deep learning features are often more discriminative than the handcrafted feature or mid-level feature. For example,~\cite{datasets:cheng2017Remote,marmanis2016deep,nogueira2017towards} demonstrated how to apply a deep learning-based architecture focus to improve the performance of remote sensing tasks. Furthermore, many efforts have been made to improve the classification accuracy by stacking, integrating or fusing multiple CNN features~\cite{zhao2016scene,li2017integrating,chaib2017deep}. In addition,~\cite{hu2015transferring} and~\cite{othman2017domain} exploit transfer learning for remote sensing image classification. Besides, there have been many efforts dedicated to classifying Hyperspectral images, such as the constrained low-rank representation based method~\cite{wang2018locality} and the adaptive subspace partition based method~\cite{wang2019hyperspectral}. Others have explored directions such as using hierarchical features~\cite{wang2017aggregating}, metric learning~\cite{cheng2018deep}, weakly supervised learning~\cite{yao2016semantic} or attention based methods~\cite{chen2018recurrent,wang2018scene}. Among these methods, learning transformation-invariant or view-invariant feature based methods often obtain better results~\cite{luus2015multiview,cheng2016learning,yu2017deep,chen2018recurrent}. 

More recently, second-order statistics features have achieved promising performance in many visual classification tasks. For example, bilinear pooling~\cite{sos:lin2015bilinear,sos:lin2017improved} collects second-order statistics of the local CNN features. However, it suffers from the high-dimensionality and the visual burstiness phenomenon. Thus, many algorithms have been proposed to reduce the feature dimensions by learning the compact feature approximation of bilinear pooling~\cite{sos:gao2016compact} or the low-rank bilinear classifier~\cite{sos:li2017factorized,sos:kong2017low}. Many algorithms have also been proposed to deal with the visual burstiness phenomenon using matrix decomposition and normalisation techniques. Examples include Improved bilinear pooling~\cite{sos:lin2017improved}, MPN-COV~\cite{sos:li2017second}, G$ ^{2} $DeNet~\cite{sos:wang2017g2denet} and Grassmann pooling~\cite{sos:wei2018grassmann}. Additionally,~\cite{he2018remote,chen2018recurrent} attempted to improve RSSC performance by stacking multi-layer covariance descriptors or introducing additional transformations.           
\section{Proposed Method} \label{method} 
\label{algorithm}
\begin{figure*}[]
	\centering
	\includegraphics[width=\textwidth]{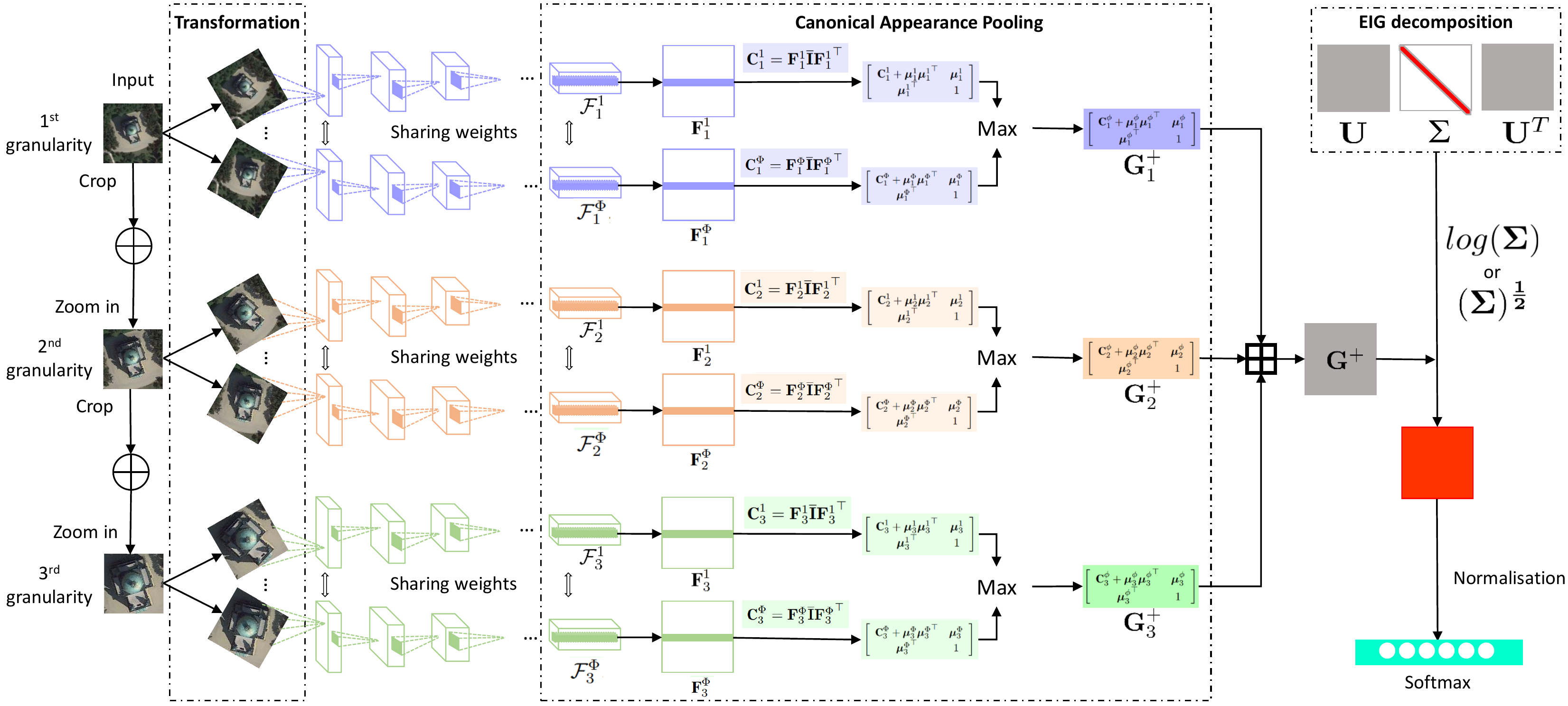}
	\caption{An overview of our MG-CAP framework. We investigate three different granularities from coarse to fine. At one specific grain level, the image is transformed according to a set of pre-defined transformation $ \Phi $. Then, the transformed image as instances will be fed into the Siamese networks for feature extraction $ \mathcal{F}_s^{\phi} $. $ \mathcal{F}_s^{\phi} $ will be transformed into a Gaussian covariance feature $ (\textbf{G}_s^{\phi})^+ $. Subsequently, we use an element-wise max operation to learn the optimal covariance feature $ \textbf{G}_s^{+} $. To capture multi-grained information, we apply an element-wise stacking operation $ \boxplus $ and average it to obtain $ \textbf{G}^+ $. The obtained $ \textbf{G}^+ $ is an SPD matrix, which can be factorised by EIG decomposition through powerful matrix normalisation methods.}
	\label{framework}
\end{figure*} 
\subsection{Overview}
The core idea of our MG-CAP model is to granularly learn multiple transformation-invariant features to reduce the large visual-semantic discrepancy in remote sensing images. Specifically, we seek a solution that deals with the challenging problems in the RSSC task without requiring detailed annotations. An illustration of the proposed MG-CAP model is shown in Fig~\ref{framework}.  

Throughout the paper, we will employ boldface lowercase letters (\textit{e.g.}, $ \textbf{v}\in\mathbb{R}^{I_1} $), boldface uppercase letters (\textit{e.g.}, $ \textbf{M}\in\mathbb{R}^{I_1\times I_2} $) and calligraphic letters (\textit{e.g.}, $ \mathcal{T}\in\mathbb{R}^{I_1\times I_2\times\dots\times I_N} $) to denote vectors, matrices and higher-order tensors, respectively. Given an input image $ \mathcal{X} \in \mathbb{R}^{H\times W\times C} $, $ H, W $ and $ C $ are the height, width and channel of the image. We first crop the raw image to produce multi-grained regional images, and then amplify them to the same size. Each regional image is transformed according to a pre-defined set of transformations. Then, we adopt Siamese style networks to extract the CNN features: $ \mathcal{F}_s \in \mathbb{R}^{H^{'}\times W^{'}\times C^{'}} $, where $ s $ is the index of granularity. Inspired by recent studies on second-order statistics~\cite{sos:acharya2018covariance,sos:ionescu2015training,sos:lin2015bilinear,sos:lin2017improved}, we flatten ordinary CNN features $ \mathcal{F}_s $ into matrices $ \textbf{F}_s\in\mathbb{R}^{H^{'}W^{'}\times C^{'}} $ that can be used to compute the covariance matrices. Subsequently, we transform the covariance matrices into the SPD form of Gaussian matrices and denote them as $ \textbf{G}_s^+\in\mathbb{R}^{(C^{'}+1)\times (C^{'}+1)} $. We learn the maximum response of SPD matrices for different granularities $ \mathcal{X}_s $, and then fuse the resulting matrices into a unique SPD matrix. After this, we use EIG function based matrix normalisations to further improve the discriminative power of features. The goal of our task is essentially to learn discriminative features containing multi-granularity information and then use it to generate a probability distribution \textbf{p} over all categories. This process can be written as:
\begin{equation}\label{eq1}
	\mathcal{X}\mapsto\textbf{G}^+\in Sym^+,\hspace{20pt}\textbf{p}(\textbf{G}^+) = f_r({\textbf{W}}\circ\textbf{G}^+)
\end{equation}

where $\mathcal{X}\mapsto\textbf{G}^+$ denotes the procedure of achieving Gaussian covariance features $\textbf{G}^+$ from an input image $\mathcal{X}$. We use $ Sym^+ $ to denote the property of the SPD matrix. It is worth mentioning that $\textbf{G}^+$ is an SPD matrix since it is the average product of the multi-grained SPD matrices. $ f_r(\cdot)$ represents the softmax layer, which maps the weighted SPD matrix $\textbf{W}\circ\textbf{G}^+$ to a feature vector and then converts it to the probabilities. $ \textbf{W} $ denotes the overall model parameters for the granular feature representation $ \textbf{G}^+ $ which can be achieved by averaging the multi-grained SPD matrices in an element-wise manner. We represent it as:
\begin{equation}\label{eq2}
	\textbf{G}^+ = \frac{1}{S}\sum\limits_{s = 1}^S {\textbf{G}_s^+}
\end{equation}
where ${\textbf{G}_s^+}$ denotes a specific SPD matrix at the corresponding granularity, which can be derived from the canonical appearance pooling layers (see the following subsection). $ S $ is the total number of granularities. Once all the canonical SPD matrices are obtained, the channel-wise average operator is applied to obtain the unique SPD matrix $\textbf{G}^+$, which incorporates information from different granularities.
\subsection{Canonical Appearance Pooling Layers}
Canonical appearance pooling layers are applied for learning the transformation-invariant feature. This kind of feature helps to deal with the problem of nuisance variations in the input image, especially those stemming from the large intra-class variations. The Multiple Instance Learning (MIL) algorithm can be applied to learn the average of the different models with the same topology. By exploiting the MIL algorithm, the learned features can be invariant to the expert-defined transformations, but the dependencies cannot be captured. 

To cope with the above problems, we need to learn the feature of transformed instances that has the highest response to classification. Notably, we utilise a Siamese style architecture to extract features of transformed images. The feature extraction process can be written as:
\begin{equation}\label{eq3}
	\mathcal{F}_s^{\phi}=f_e\left(\phi \left({\mathcal{X}_s}\right)\right),
\end{equation}
where $ \phi  \in \Phi $ is a set of pre-defined transformations. In this paper, we only consider rotation transformations that can be derived from:$\phi_{r}=\frac{360^{\circ}}{\text{dim}(\Phi)}$, where \text{dim}($\cdot$) denotes the length of a transform set. $ \phi ({\mathcal{X}_s}) $ denotes the transformed images, and $ f_e(\cdot) $ indicates the feature extraction process with using the standard deep CNN architecture.  

The above procedure is easy to implement but is invariant under certain transformations. Specifically, we adopt the maximum operator to learn the optimal feature representation from the transformed instances. We formulate it as:                  
\begin{equation}\label{eq4}
	\mathcal{F}_s^{\phi}\mapsto(\textbf{G}_s^{\phi})^+, \hspace{20pt}\textbf{G}_s^+ = \mathop {\max }\limits_{\phi  \in \Phi } f_c\left((\textbf{G}_s^{\phi})^+\right),
\end{equation}
where $ \mathcal{F}_s^{\phi}\mapsto(\textbf{G}_s^{\phi})^+ $ is a procedure that transforms CNN features into Gaussian covariance matrices. $ f_c(\cdot) $ is adopted to learn the optimal second-order feature from the accumulated covariance matrices $ (\textbf{G}_s^{\phi})^+ $. In this way, the generated feature $\textbf{G}_s^+$ can be seen as a new feature with the transformation-invariant property.

The most direct way to achieve the above objective is to use the MIL algorithm. However, it has a tremendous demand for computational capacity and capability. Furthermore, the MIL algorithm neglects feature dependencies because it treats the instance features independently. To this end, we adopt the Siamese style architecture~\cite{chopra2005learning}, which allows weights to be shared among the different instance features. The obtained feature is not only transformation invariant but also maintains the same model parameters as when using an individual instance. To incorporate the multi-grained information, we deploy it in multi-stream fashion but prevent weights from being shared between neighbouring granularities (\textit{i.e.,} each granularity is independent).

Due to the vectorisation of the spatial locations of features, traditional deep learning frameworks can only capture first-order information. This first-order feature preserves the invariance of CNN features and achieves competitive performance in many image classification tasks. However, the spatial information of the image can influence the final categorisation performance (\textit{e.g.,} in our remote sensing scene recognition scenario) and needs to be maintained selectively. To overcome this problem, we regard CNN features as matrices and then transform them into covariance matrices, which can effectively preserve the spatial structures of image regions.

On a specific grain level, the transformation-invariant covariance feature $\textbf{G}_s^+$ can be obtained through Eq.~\ref{eq4}. Specifically, CNN features $\mathcal{F}_s^{\phi}$ need to be converted into covariance matrix ($\textbf{G}_s^{\phi})^+$ before the pooling operation. CNN features can be expressed in a matrix form $ \mathbf{F}_s^{\phi}=[\textbf{f}_1,\textbf{f}_2,...,\textbf{f}_N] $ by flattening the spatial structure of $\mathcal{F}_s^{\phi}$, where $ \textbf{f}_i \in \mathbb{R}^{C^{'}} $ and $ N = H^{'}\times W^{'} $. The computation of the covariance matrix can be seen as the compact summarisation of the second-order information of $\mathcal{F}_s^{\phi} $, which is given by:         
\begin{equation}\label{eq5}
	{\textbf{C}_{s}^{\phi}} = {\mathbf{F}_{s}^{\phi}}\overline{\mathbf{I}}{{\mathbf{F}_{s}^{\phi}}}^{\top},
\end{equation}
where $\textbf{C}_s^{\phi}\in\mathbb{R}^{C^{'}\times C^{'}}$. $\overline{\mathbf{I}}$ can be calculated as:$ \overline{\mathbf{I}}=\frac{1}{N}(\mathbf{I}-\frac{1}{N}\textbf{11}^{\top}) $ with $\mathbf{I}\in \mathbb{R}^{N\times N}$ and $\mathbf{1}=[1,\dots,1]^{\top}$. $ N $ denotes the vector dimensions. 

The obtained $ \textbf{C}_s $ encodes the second-order statistics of local CNN features. Especially, covariance matrices $ \textbf{C}_s $ are SPD matrices when their components are linearly independent in the corresponding vector feature spaces $ \mathbf{F}_s^{\phi} $ and the spatial number $ N $ is greater than C$ ^{'} $. As suggested by~\cite{sos:acharya2018covariance,sos:huang2015log}, the SPD form of the Gaussian matrix is usually superior to the standard covariance matrix in classification tasks since it simultaneously incorporates the first-order and the second-order information of CNN features. The Gaussian covariance matrices can be obtained by transforming $ \textbf{C}_s^{\phi} $ into a single Gaussian model $ \mathcal{N}(\boldsymbol{\mu},\textbf{C}_s^{\phi}) $. We write this as:
\begin{equation}\label{eq7}
	\textbf{G}_s^{\phi} = \left[ {\begin{array}{*{20}{c}}
			{\textbf{C}_s^{\phi} + \boldsymbol{\mu} {{\boldsymbol{\mu}}^{\top}}}&\boldsymbol{\mu} \\
			{{{}\boldsymbol{\mu}^{\top}}}&1
	\end{array}} \right],
\end{equation}
where $\boldsymbol{\mu}=\sum_{n=1}^{N}\textbf{f}_{n}$. The dimension of the Gaussian covariance matrix $ \textbf{G}_s $ is $ (C^{'}+1)\times( C^{'}+1)$. The elements of the obtained covariance matrix naturally reside on the Riemannian manifold of the SPD matrix. A directly flattening operation will damage the geometry of the Riemannian manifold in $ \textbf{G}_s^{\phi} $. Instead, we apply the logarithmic operation for flattening the spatial structure of the Riemannian manifold so that all of the distance measurements in the Euclidean space can be adopted. Besides, to maintain the singularity of $ \textbf{G}_s^{\phi} $, a small ridge is introduced and added to the Gaussian covariance matrix $ \textbf{G}_s $:     
\begin{equation}\label{eq8}
	(\textbf{G}_s^{\phi})^+ = \textbf{G}_s^{\phi} + \lambda \text{trace}(\textbf{G}_s^{\phi})\textbf{I}_{g},
\end{equation}
where $ \lambda $ is a hyper-parameter and $ \textbf{I}_{g}\in \mathbb{R}^{(C^{'}+1)\times( C^{'}+1)} $ denotes the identity matrix. Specifically, it can be seen as a regularisation term to convert the resulting symmetric positive semi-definite matrix into a symmetric positive definite matrix. 

Encoding second-order statistics brings valuable regional information when compared with the first-order features.  As illustrated in Eq.\ref{eq5}-Eq.~\ref{eq8}, we have shown the flexibility of learning second-order features through the standard covariance matrix or Gaussian covariance matrix. The most commonly adopted feature fusion method is to concatenate the different features along with the channel dimension. Nevertheless, this creates a high-dimensional feature map and leads to an exponential increase in computational time. Instead of a concatenation, we propose to fuse features by stacking and averaging the SPD matrices ( see the latter part of Eq.~\ref{eq2} ). Notably, this enables us to capture the multi-grained information while preserving the over-expansion of the feature dimension of $ \textbf{G}^+ $.    
\subsection{EIG-decomposition Layers}
The obtained SPD matrix $ \textbf{G}^+ $ can be directly fed into the classifier. However, the feature is more discriminative if we appropriately exploit the geometric structure of the SPD manifold. This motivates us to apply the EIG-decomposition function for factorising the SPD matrix, especially, in the nonlinear scenario. The EIG-decomposition function offers an effective way to scale the spectrum of the SPD matrix using the appropriate normalisation methods. Primarily, the matrix power is represented by the power of eigenvalues. We express the process of EIG-decomposition as:          
\begin{equation}\label{eq9}
	(\textbf{G}^ +)_k = f_d^{(k)}\left((\textbf{G}^ +)_{k-1}\right) = {\textbf{U}_{k - 1}}\textbf{F}({\boldsymbol{\Sigma}} _{k - 1}) \textbf{U}_{k - 1}^{\top},
\end{equation}
where $ \textbf{F}({\boldsymbol{\Sigma}} _{k - 1}) $ is the normalised diagonal matrix of eigenvalues and can be denoted as follows:
\begin{equation}\label{eq10}
	\textbf{F}({\boldsymbol{\Sigma}} _{k - 1})= \left\{ \begin{array}{l}
	\text{diag}\left(\text{log}(\nu_1),\cdots,\text{log}(\nu_c)\right) ;\\
	\text{diag}\left((\nu_1)^{\frac{1}{2}},\cdots,(\nu_c)^{\frac{1}{2}}\right).
	\end{array} \right.
\end{equation}
Here $\text{diag}(\cdot)$ is a diagonal operation of the matrix, and $\text{log}(\nu_i) $ is the logarithm of eigenvalues $ \nu_i$ with $ i=1,...,c $, where $ c=C^{'}+1 $, arranged in non-increasing order. 

As shown in Eq.~\ref{eq10}, we present two approaches to normalise eigenvalues. For SPD matrices, the natural choice is to compute the logarithm of eigenvalues since it succeeds in endowing the Riemannian manifold of SPD matrices with a Lie group structure~\cite{sos:acharya2018covariance}. Consequently, the flattened Riemannian space allows the computational operations in Euclidean space to be applied in Log-Euclidean space. Although a point in the tangent space can approximate the flattened SPD manifold locally, the logarithm of the eigenvalues matrix (Log-E) is often numerically unstable in the non-linear scenario. The square root of the eigenvalue matrix (Sqrt-E), as an alternative stable solution, has attracted increasing attention.

The Log-E metric requires eigenvalues to be strictly positive and it considerably changes the magnitudes of eigenvalues. Specifically, it often overstretches the small eigenvalues and reverses the order of eigenvalues, which influences the importance of the inherent linear relationship. To avoid these problems, we adopt the rectification function introduced in~\cite{sos:acharya2018covariance} and rewrite it as:
\begin{equation}\label{eq11}
	\textbf{R} = \max (\varepsilon \textbf{I},\boldsymbol{\Sigma}_{k - 1}),
\end{equation}
where $ \varepsilon $ is a threshold and $ \textbf{I} $ is an identity matrix. To prevent elements of eigenvalues from being close to non-positive ones, we replace $\nu_i$ by $\textbf{R}(i, i)$ in the sequel. The above function is similar to the ReLU activation function in neural networks~\cite{glorot2011deep} and can be viewed as a non-linear rectification function. However, it is more robust and ideal in our scenarios since ReLU usually yields sparsity~\cite{glorot2011deep}. Specifically, the diagonal elements can be defined as:
\begin{equation}\label{eq12}
	\textbf{R}(i,i) = \left\{ \begin{array}{l}
		\boldsymbol{\Sigma}_{k - 1}(i,i),\hspace{32pt}\boldsymbol{\Sigma}_{k - 1}(i,i) > \varepsilon ;\\
		\varepsilon ,\hspace{70pt}\boldsymbol{\Sigma}_{k - 1}(i,i) \le \varepsilon,
	\end{array} \right.
\end{equation} 
where $ \boldsymbol{\Sigma}_{k - 1} = \text{diag}(\nu_1,\cdots,\nu_c) $ and can be obtained by the standard EIG function as follows:
\begin{equation}\label{eq13}
	(\textbf{G}^ +)_{k-1}={\textbf{U}_{k - 1}}{\boldsymbol{\Sigma}} _{k - 1} \textbf{U}_{k - 1}^{\top}.
\end{equation}

The above rectification layer is devised for the Log-E metric to ensure the normalised eigenvalues are positive real numbers and then improve the numerical stability. This function is not specified to the Sqrt-E metric because the square root normalisation allows non-negative eigenvalues. A similar rectification function was introduced in~\cite{sos:acharya2018covariance}. However, it was previously incorporated into an extra EIG-decomposition function, leading to significant demands in computational costs and making it time-consuming. In contrast, we introduce a proxy parameter $\textbf{R}(i, i)$ to allow the EIG-decomposition function to be called only once. 
\subsection{Back-propagation}\label{bp}
Deep learning relies heavily on efficient gradient computation algorithms for back-propagation. However, existing methods~\cite{sos:lin2017improved,sos:li2017second} usually compute eigenvalues in CPUs since the EIG-decomposition function is not well supported on the CUDA platform. Specifically, the gradient of the EIG-decomposition function approaches infinity when the input matrices are degenerate. This implies that one or more of its normalised eigenvalues may be identical. Namely, the corresponding eigenvectors can be arbitrary in this situation. To circumvent this problem, we replace the infinite gradient values with 0, which effectively prevents the gradient computation from being interrupted during back-propagation.    

To demonstrate the back-propagation of our algorithm, we adopt the method introduced in~\cite{sos:ionescu2015training} which computes the gradient of the general matrix by establishing the corresponding chain rule with first-order Taylor expansion and approximation. Compared with the back-propagation for the standard EIG-decomposition function, we also provide the gradient for the normalisation function and rectification function. Given the final loss function $l$, the gradients for the classification layer in Eq.~\ref{eq1} can be calculated as $L_{c}=l\circ f_{r}$. Let $ L^k =L_c\circ f_d^{(l)}\circ f_d^{(l-1)}\circ,...,\circ f_d^{(1)}$ denote the corresponding gradient in the $k$-th layer of the EIG-decomposition function. The chain rule can be expressed as:
\begin{equation}\label{eq15}
	\frac{{\partial {L^{(k)}}\left({{({\textbf{G}^ + })}_{k - 1}},y\right)}}{{\partial {{({\textbf{G}^ + })}_{k - 1}}}} = \frac{{\partial {L^{(k + 1)}}\left({{({\textbf{G}^ + })}_k},y\right)}}{{\partial {{({\textbf{G}^ + })}_k}}}\frac{{\partial {f_d^{(k)}}\left({{({\textbf{G}^ + })}_{k - 1}}\right)}}{{\partial {{({\textbf{G}^ + })}_{k - 1}}}},
\end{equation}       
where $y$ is the output of the classification layer. $ (\textbf{G}^ +)_k = f_d^{(k)}\left((\textbf{G}^ +)_{k-1}\right) $ was previously introduced in Eq.~\ref{eq9}. Suppose $\mathscr{F}$ is a function that describes the variations of the present layer to the previous layer in the EIG-decomposition function, which is written as: $ d(\textbf{G}^ +)_k = \mathscr{F}(d(\textbf{G}^ +)_{k-1})$. The chain rule becomes:
\begin{equation}\label{eq16}
	\frac{{\partial {L^{(k)}}\left({{({\textbf{G}^ + })}_{k - 1}},y\right)}}{{\partial {{({\textbf{G}^ + })}_{k - 1}}}} = \mathscr{F}^*\left( {\frac{{\partial {L^{(k + 1)}}\left({{({\textbf{G}^ + })}_k},y\right)}}{{\partial {{({\textbf{G}^ + })}_k}}}} \right),
\end{equation}   
where $ \mathscr{F}^* $ is a non-linear adjoint operator of $ \mathscr{F} $ ($ i.e. $, $ \textbf{A}:\mathscr{F}(\textbf{B}) = \mathscr{F}^*(\textbf{A}):\textbf{B}) $. Specifically, $:$ denotes the matrix inner product in the Euclidean $ vec^{'} $d matrix space, which has the property of colon-product and can be written as $ \textbf{A}:\textbf{B} = \sum\limits_{i,j}\textbf{A}_{ij}\textbf{B}_{ij}=\text{trace}({\textbf{A}^{\top}}\textbf{B}) $. As all of the operations rely on EIG-decomposition functions ($i.e.,$ Eq.~\ref{eq13}), we introduce a virtual operation ($ i.e., k^{'}$ layer ). Then, the updated chain rule that based on Eq.~\ref{eq16} can be written as:
\begin{equation}\label{eq17}
	\begin{split}
		&\frac{{\partial {L^{(k)}}({{({\textbf{G}^ + })}_{k - 1}},y)}}{{\partial {{({\textbf{G}^ + })}_{k - 1}}}}:d{({\textbf{G}^ + })_{k - 1}}\\
		=& \mathscr{F}^{*}\left(\left( {\frac{{\partial {L^{({k^{'}})}}}}{{\partial {\textbf{U}_{k - 1}}}}} \right) + \left( {\frac{{\partial {L^{({k^{'}})}}}}{{\partial {\boldsymbol{\Sigma}_{k - 1}}}}} \right)\right):d{({\textbf{G}^ + })_{k - 1}}\\
		=& \frac{{\partial {L^{({k^{'}})}}}}{{\partial {\textbf{U}_{k - 1}}}}:\mathscr{F}(d{({\textbf{G}^ + })_{k - 1}}) + \frac{{\partial {L^{({k^{'}})}}}}{{\partial {\boldsymbol{\Sigma}_{k - 1}}}}\mathscr{F}(d{({\textbf{G}^ + })_{k - 1}})\\
		=& \frac{{\partial {L^{({k^{'}})}}}}{{\partial {\textbf{U}}}}:d{\textbf{U}} + \frac{{\partial {L^{({k^{'}})}}}}{{\partial {\boldsymbol{\Sigma}}}}:d{\boldsymbol{\Sigma}}.
	\end{split}
\end{equation}   
Note that we simply drop the subscripts of $d{\textbf{U}_{k - 1}}$ and $d{\boldsymbol{\Sigma}_{k - 1}}$ in the last line to improve the readability. Both $d{\textbf{U}}$ and $d{\boldsymbol{\Sigma}}$ are derived from the variation of the EIG function:
\begin{equation}\label{eq18} 
	d{({\textbf{G}^ + })_{k - 1}} = d{\textbf{U}}{\boldsymbol{\Sigma}}\textbf{U}^{\top} + {\textbf{U}}d{\boldsymbol{\Sigma}}\textbf{U}^{\top} + {\textbf{U}}{\boldsymbol{\Sigma}}d\textbf{U}^{\top},
\end{equation}
After some rearrangements, $d{\textbf{U}}$ and $d{\boldsymbol{\Sigma}}$ can be denoted as:
\begin{equation}\label{eq19} 
	\begin{split}
		&d{\textbf{U}} = {\textbf{U}}({\textbf{Q}^{\top}} \odot {(\textbf{U}^{\top}d{({\textbf{G}^ + })_{k - 1}}{\textbf{U}})})\\
		&d{\boldsymbol{\Sigma}} = {(\textbf{U}^{\top}d{({\textbf{G}^ + })_{k - 1}}{\textbf{U}})_{diag}}
	\end{split}
\end{equation}
where $\odot$ denotes the Hadamard product of the matrix. Besides, ${\textbf{M}_{sym}} = \frac{1}{2}(\textbf{M} + {\textbf{M}^{\top}})$ and $ \textbf{M}_{diag} $ is $ \textbf{M} $ with all off-diagonal elements set to 0. $\textbf{Q}$ can be achieved by:
\begin{equation}\label{eq20} 
	\textbf{Q}(i,j) = \left\{ \begin{array}{l}
		\frac{1}{{{\nu _i} - {\nu _j}}},\hspace{50pt}i \ne j;\\
		0,\hspace{70pt}i = j.
	\end{array} \right.
\end{equation}

We refer readers to~\cite{sos:ionescu2015training} for more details on deriving Eq.~\ref{eq19}. The specific partial derivatives of the loss function can be derived by plugging Eq.~\ref{eq19} into Eq.~\ref{eq17}. Moreover, the properties of the matrix inner product, previously introduced can naturally be denoted as:
\begin{equation}\label{eq21}
	\begin{split}
		&\frac{{\partial {L^{(k)}}}}{{\partial {{({\textbf{G}^ + })}_{k - 1}}}}
		=\frac{{\partial {L^{(k)}}\left({{({\textbf{G}^ + })}_{k - 1}},y\right)}}{{\partial {{({\textbf{G}^ + })}_{k - 1}}}}\\
		=&{\textbf{U}}\left( {\left( {{\textbf{Q}^{\top}} \odot \left( {{\textbf{U}}^{\top}\frac{{\partial {L^{({k^{'}})}}}}{{\partial {\textbf{U}}}}} \right)} \right) + {{\left( {\frac{{\partial {L^{({k^{'}})}}}}{{\partial {\boldsymbol{\Sigma}}}}} \right)}_{diag}}} \right){\textbf{U}}^{\top}
	\end{split},
\end{equation}
where $\frac{\partial {L^{({k^{'}})}}}{{\partial {\textbf{U}}}}$ and $\frac{\partial {L^{({k^{'}})}}}{{\partial {\boldsymbol{\Sigma}}}}$ can be calculated by employing a similar strategy as $\frac{{\partial {L^{(k)}}}}{{\partial {{({\textbf{G}^ + })}_{k - 1}}}}$ in Eq.~\ref{eq17}. Then, we can derive the partial derivatives of $ \frac{{\partial {L^{({k^{'}})}}}}{{\partial \textbf{U}}} $ and $ \frac{{\partial {L^{({k^{'}})}}}}{{\partial \boldsymbol{\Sigma}}}$ as: 
\begin{equation}\label{eq22}
	d{({\textbf{G}^ + })_k} = 2{(d\textbf{U}g(\boldsymbol{\Sigma}){\textbf{U}^{\top}})_{sym}} + \textbf{U}{g^{'}}(\boldsymbol{\Sigma})d\boldsymbol{\Sigma}{\textbf{U}^{\top}}.
\end{equation} 
Following the chain rule introduced in Eq.~\ref{eq17}, the derivative of $ \frac{{\partial {L^{({k^{'}})}}}}{{\partial \textbf{U}}} $ and $ \frac{{\partial {L^{({k^{'}})}}}}{{\partial \boldsymbol{\Sigma}}}$ can be obtained by:
\begin{equation}\label{eq23}
	\begin{split}
		\frac{{\partial {L^{({k^{'}})}}}}{{\partial \textbf{U}}} &= 2{\left( {\frac{{\partial {L^{(k + 1)}}}}{{\partial {{({\textbf{G}^ + })}_k}}}} \right)_{sym}}\textbf{U}g(\boldsymbol{\Sigma}) \\
		\frac{{\partial {L^{({k^{'}})}}}}{{\partial \boldsymbol{\Sigma}}} &= {g^{'}}(\boldsymbol{\Sigma}){\textbf{U}^{\top}}\left( {\frac{{\partial {L^{(k + 1)}}}}{{\partial {{({\textbf{G}^ + })}_k}}}} \right)\textbf{U}.
	\end{split}
\end{equation}
The partial derivative of $ \frac{{\partial {L^{({k^{'}})}}}}{{\partial \textbf{U}}} $ and $ \frac{{\partial {L^{({k^{'}})}}}}{{\partial \boldsymbol{\Sigma}}}$ in Log-E form can be written as: 
\begin{equation}\label{eq24}
	\begin{split}
		\frac{{\partial {L^{({k^{'}})}}}}{{\partial \textbf{U}}}&= 2{\left( {\frac{{\partial {L^{(k + 1)}}}}{{\partial {{({\textbf{G}^ + })}_k}}}} \right)_{sym}}\textbf{U}\log (\textbf{R})\\
		\frac{{\partial {L^{({k^{'}})}}}}{{\partial \boldsymbol{\Sigma}}}&= \text{diag}(\nu _1^{ - 1},...,\nu _c^{ - 1}){\textbf{U}^{\top}}\left(\frac{{\partial {L^{(k + 1)}}}}{{\partial {{({\textbf{G}^ + })}_k}}}\right)\textbf{U}.
	\end{split}
\end{equation}
Similarly, the partial derivative in Sqrt-E form is obtained by:
\begin{equation}\label{eq25}
	\begin{split}
		\frac{{\partial {L^{({k^{'}})}}}}{{\partial \textbf{U}}}&= 2{\left( {\frac{{\partial {L^{(k + 1)}}}}{{\partial {{({\textbf{G}^ + })}_k}}}} \right)_{sym}}\textbf{U}(\textbf{R})^\frac{1}{2}\\
		\frac{{\partial {L^{({k^{'}})}}}}{{\partial \boldsymbol{\Sigma}}}&= \frac{1}{2\sqrt{\nu_1}}\left(\text{diag}(\nu _1^{ -\frac{1}{2}},...,\nu _c^{ - \frac{1}{2}}){\textbf{U}^{\top}}\left(\frac{{\partial {L^{(k + 1)}}}}{{\partial {{({\textbf{G}^ + })}_k}}}\right)\textbf{U}\right)\\&-\text{diag}\left(\frac{1}{2\sqrt{\nu_1}}\text{trace}\left((\textbf{G}^ +)_{k}\frac{{\partial {L^{(k + 1)}}}}{{\partial {{({\textbf{G}^ + })}_k}}}\right),0,\dots,0\right),
	\end{split}
\end{equation}
where $ \textbf{R}$ is the resulting matrix introduced in Eq.\ref{eq11}. Specifically, $ g^{'}(\boldsymbol{\Sigma}) $ in $ {g^{'}}(\boldsymbol{\Sigma}){\textbf{U}^{\top}}\left( {\frac{{\partial {L^{(k + 1)}}}}{{\partial {{({\textbf{G}^ + })}_k}}}} \right)\textbf{U} $ will be replaced by $ {g^{'}}(\textbf{R}){\textbf{U}^{\top}}\left( {\frac{{\partial {L^{(k + 1)}}}}{{\partial {{({\textbf{G}^ + })}_k}}}} \right)\textbf{U} $ in the sequel. The corresponding gradient of $ \textbf{R} $ can be calculated by:
\begin{equation}\label{eq26}
	\textbf{R}(i,i) = \left\{ \begin{array}{l}
		1,\hspace{30pt}\boldsymbol{\Sigma}_{k - 1}(i,i) > \varepsilon ;\\
		0 ,\hspace{30pt}\boldsymbol{\Sigma}_{k - 1}(i,i) \le \varepsilon.
	\end{array} \right.
\end{equation} 
Once the partial derivatives of $ \frac{{\partial {L^{({k^{'}})}}}}{{\partial \textbf{U}}} $ and $ \frac{{\partial {L^{({k^{'}})}}}}{{\partial \boldsymbol{\Sigma}}}$ have been obtained, they can be plugged into Eq.\ref{eq21}, resulting in the back-propagation of the Riemannian SPD matrices with logarithm normalisation and more robust square-root normalisation.

When the partial derivative of $ \frac{{\partial {L^{(k)}}}}{{\partial {{({\textbf{G}^ + })}_{k - 1}}}} $ has been obtained, it can be used to compute the gradient for learning canonical transformation. We introduce $ \nabla f_c\left((\textbf{G}_s^{\phi})^+\right) $ to denote the gradient of the feature $f_c\left((\textbf{G}_s^{\phi})^+\right)$. Then, we can derive the gradient of $ \frac{{d{\textbf{G}_s^+}}}{{d{f_c\left((\textbf{G}_s^{\phi})^+\right)}}}$ as:
\begin{equation}\label{eq28}
	\frac{{d{\textbf{G}_s^+}}}{{d{f_c\left((\textbf{G}_s^{\phi})^+\right)}}} = \nabla f_c\left((\textbf{G}_s^{\phi})^+\right),
\end{equation}
where $ \phi=\underset{\phi  \in \Phi} {\mathrm{argmax}} ~f_c\left((\textbf{G}_s^{\phi})^+\right) $ is the optimal appearance $ \phi $ with respect to input $ \mathcal{X}_s $ at a specific granularity.  

Finally, we can derive the gradient of the loss function for matrix $ \textbf{F}_s^{\Phi} $ that is reshaped by CNN features in Eq~\ref{eq5}. Specifically, it can be expressed as:
\begin{equation}\label{eq27}
	\frac{{\partial {L^{(k)}}}}{\partial \textbf{F}_s^{\Phi}}=\left( \frac{{\partial {L^{(k)}}}}{\partial \textbf{F}_s^{\Phi}}+\left(\frac{{\partial {L^{(k)}}}}{\partial {\textbf{F}_s^{\Phi}}}\right)^{\top}\right)\overline{\textbf{I}}\textbf{F}_s^{\Phi}.
\end{equation}
The above formulations have shown the back-propagation of our framework in detail. For clarity, we derive the gradient of our framework for the EIG-decomposition layers, SPD matrices layers and canonical appearance pooling layers in sequence. The gradient of an integrated framework can be computed by cascading different granularities together since our MG-CAP architecture is fully differentiable.             
\section{Experiments}\label{experiments}
\subsection{Experimental Datasets}
\begin{table}[!htbp]
	\caption{The statistics of datasets used for remote sensing scene categorisation in this paper}
	\begin{tabular}{c|c|c|c}
		\hline
		Datasets      & \ No. Images & \ No. Class & \ No. Class Images \\ \hline
		NWPU-RESISC45~\cite{datasets:cheng2017Remote} & 31,500        & 45           & 700                          \\
		AID~\cite{datasets:xia2016AID}           & 10,000        & 30           & 220$\sim$420                 \\
		UC Merced~\cite{yang2010bag}     & 2,100         & 21           & 100                       \\ \hline   
	\end{tabular}\label{datasets}
\end{table}
We conduct our experiments on three well-known datasets that are widely used for evaluating the RSSC task. These include NWPU-RESISC45~\cite{datasets:cheng2017Remote}, Aerial Image Dataset (AID)~\cite{datasets:xia2016AID} and UC Merced~\cite{yang2010bag}. In addition to the primary indices summarised in TABLE~\ref{datasets},  the spatial resolution is another important factor for the RSSC task. For example, it is 0.3m in the UC Merced~\cite{yang2010bag} dataset, 0.5m to 8m in AID~\cite{datasets:xia2016AID} and 0.2m to 30m in NWPU-RESISC45~\cite{datasets:cheng2017Remote}.
\subsection{Implementation Details}
For fair comparison, we implement our framework using the VGG-D~\cite{simonyan2014very} architecture, pre-trained on the large-scale ImageNet dataset. During training, data augmentation techniques are adopted to avoid overfitting. These include randomly cropping 224 $ \times $ 224 patches from 256 $ \times $ 256 images, followed by horizontal flipping. Then, we transform the image and pad it to 317 $ \times $ 317 pixels with zero-padding (Note: only the position of the original pixels is changed). Subsequently, we employ bilinear interpolation to resize the transformed images into 224 $\times$ 224, which is convenient for feeding them into the Siamese architecture for feature extraction. We retain model parameters before the last non-activated convolutional features of VGG-D~\cite{simonyan2014very} (\textit{i.e.,} conv5\_3). 

We initially train the classification layer using a learning rate of 0.1 and then fine-tune the entire network with a small learning rate of $ 10^{-3} $. The learning rate is annealed by 0.15 every 30 epochs during the warm-up stage and then decayed after every 3 epochs during the fine-tuning stage. The batch size is 12 for the experiments of 3 granularities with 12 different transformations. The weight decay rate is $ 5 \times 10^{-4} $. The framework is optimised using the momentum optimiser with a constant momentum factor of $0.9$. It is worth noting that we randomly split the datasets for training and testing ten times. Meanwhile, we report the mean and standard deviation of obtained classification results. We empirically set $ \lambda $ (in Eq.~\ref{eq8}) as $ 1\times10^{-4} $. The threshold parameter $ \varepsilon $ in Eq.~\ref{eq11} is used to rectify the value of eigenvalues. It was originally introduced in~\cite{sos:acharya2018covariance} for the Log-E metric. We employ it for both Sqrt-E and Log-E in our experiments, where is used to clip the eigenvalue into $ [1\times10^{-5},1\times10^{5}] $. Besides, the framework is implemented using the GPU version of TensorFlow 1.0.      
\subsection{Experimental Results and Comparisons}
% \usepackage{multirow}
\begin{table}[]
	\centering
	\caption{Comparison of classification results (\%) achieved by our MG-CAP and previous works on NWPU-RESISC45~\cite{datasets:cheng2017Remote}.}
	\scalebox{0.92}{
		\begin{tabular}{cccc}
			\toprule
			\multirow{2}{*}{}                                                                         & \multirow{2}{*}{Method} & \multicolumn{2}{c}{NWPU-RESISC45} \\\cline{3-4} 
			&                         & T.R.=10\%       & T.R.=20\%       \\\hline
			\multirow{3}{*}{Handcrafted Feature}                                                        & GIST~\cite{datasets:cheng2017Remote}                    & 15.90$ \pm $0.23      & 17.88$ \pm $0.22      \\
			& LBP~\cite{datasets:cheng2017Remote}                      & 19.20$ \pm $0.41      & 21.74$ \pm $0.18      \\
			& Colour Histogram~\cite{datasets:cheng2017Remote}        & 24.84$ \pm $0.22      & 27.52$ \pm $0.14      \\\hline
			\multirow{3}{*}{\begin{tabular}[]{@{}c@{}}Unsupervised Feature\\ Learning\end{tabular}}
	 		& BoVW+SPM~\cite{datasets:cheng2017Remote}                 & 27.83$ \pm $0.61      & 32.96$ \pm $0.47      \\
			& LLC~\cite{datasets:cheng2017Remote}                      & 38.81$ \pm $0.23      & 40.03$ \pm $0.34      \\
			& BoVW~\cite{datasets:cheng2017Remote}                     & 41.72$ \pm $0.21      & 44.97$ \pm $0.28      \\\hline
			\multirow{10}{*}{Deep Learning}                                                            & AlexNet~\cite{cheng2018deep}                  & 76.69$ \pm $0.21      & 79.85$ \pm $0.13      \\
			& GoogLeNet~\cite{cheng2018deep}                & 76.47$ \pm $0.18      & 79.79$ \pm $0.15      \\
			& VGG-D~\cite{cheng2018deep}                    & 76.19$ \pm $0.38      & 78.48$ \pm $0.26      \\
			& AlexNet+BoVW~\cite{cheng2017remote_bag}            & 55.22$ \pm $0.39      & 59.22$ \pm $0.18      \\
			& GoogLeNet+BoVW~\cite{cheng2017remote_bag}            & 78.92$ \pm $0.17     & 80.97$ \pm $0.17      \\
			& VGG-D+BoVW~\cite{cheng2017remote_bag}              & 82.65$ \pm $0.31      & 84.32$ \pm $0.17    \\
			\cline{2-4}
			&\textbf{MG-CAP (Bilinear)}                                      &{89.42}$\pm${0.19}                     &{91.72}$\pm${0.16}\\
			&\textbf{MG-CAP (Log-E)}                                      &{88.35}$\pm${0.23}                     &{90.94}$\pm${0.20}\\
			&\textbf{MG-CAP (Sqrt-E)} &\textbf{90.83}$\pm$\textbf{0.12} &\textbf{92.95}$\pm$\textbf{0.11}\\\bottomrule
		\end{tabular}\label{results_on_NU45}}
\end{table}
We compare our framework with several benchmark methods on the most challenging datasets~\cite{datasets:cheng2017Remote}. As shown in TABLE~\ref{results_on_NU45}, the Colour histograms method achieves the best classification results among the listed handcrafted feature-based methods. Specifically, the Colour histogram feature performs better than the LBP feature and the global GIST feature under the two different datasets partitions. Furthermore, unsupervised feature learning-based methods achieve higher accuracy than all of the listed handcrafted feature-based methods. An algorithm combining BoVW and SPM was proposed to incorporate more spatial information from images, but it only achieves accuracies of 27.83\% and 32.96\%~\cite{datasets:cheng2017Remote}. The LLC is slightly better than BoVW+SPM, but still falls short when compared with the BoVW algorithm (\textit{i.e.,} about 3\% and 5\% differences). Deep learning-based methods demonstrate their superior performance and overshadow both handcrafted and unsupervised based feature learning methods. To be precise, with a linear SVM classifier, transferred neural networks~\cite{cheng2018deep} achieve an accuracy of about 76\% for the 10\% training split, while the accuracy is further increased by about 3\% for the 20\% training ratio. Furthermore, the combination of VGG-D and BoVW achieves the best performance among all registered methods. However, the combination of AlexNet and BoVW only produces accuracies of 55.22\%$\pm$0.39 and 59.22\%$\pm$0.18~\cite{cheng2017remote_bag}, which is surprisingly lower than other deep learning-based algorithms and even lower than the transferred AlexNet~\cite{cheng2018deep}.
\begin{figure}[]
	\centering
	\includegraphics[width=0.48\textwidth]{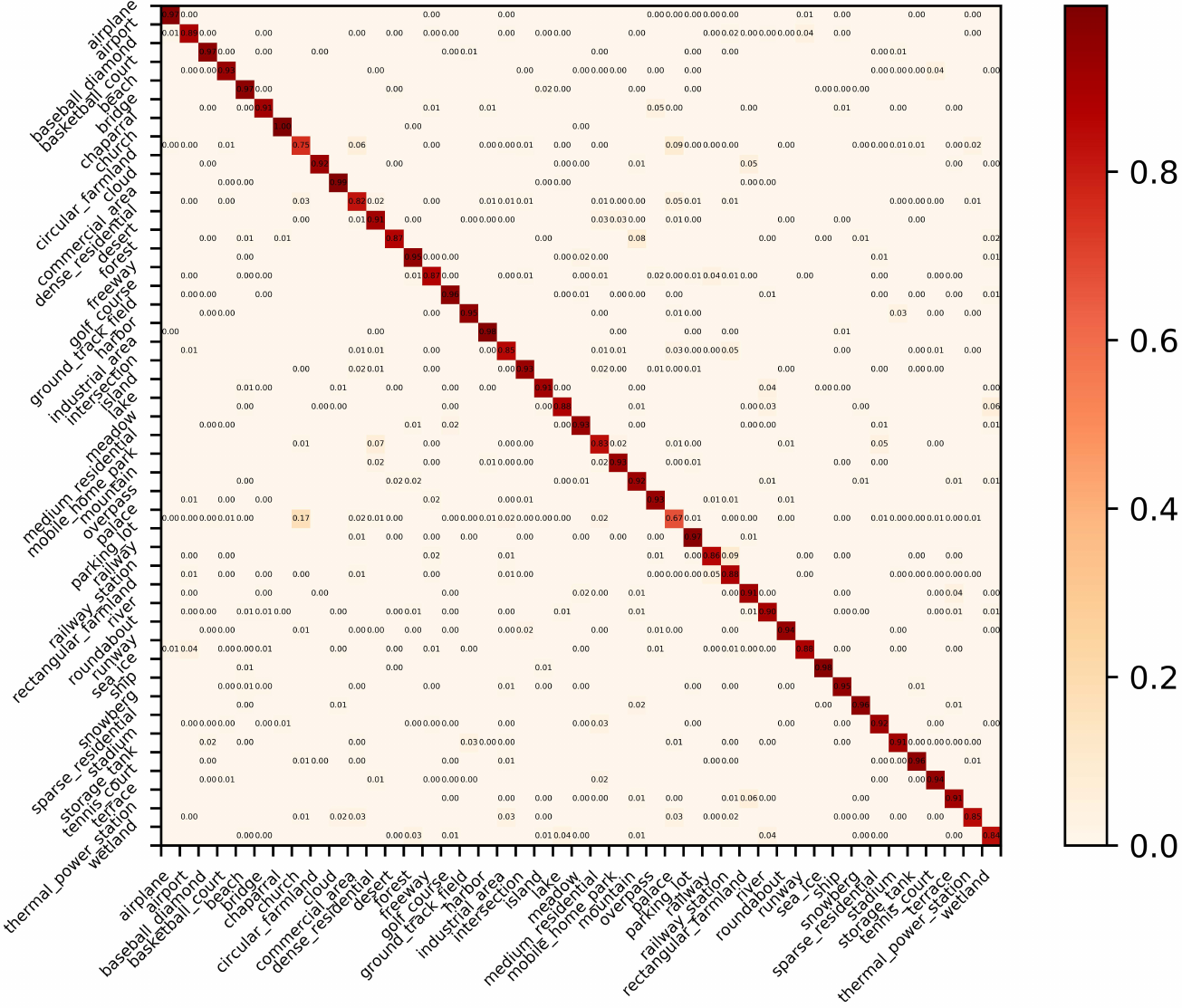}
	\caption{Confusion matrix for NWPU-RESISC45~\cite{datasets:cheng2017Remote} with 10\% training data.}
	\label{mg_cap:nwpu_cm}
\end{figure} 

We provide three different variations based on our MG-CAP framework, which include the original bilinear pooling, the logarithm of the eigenvalue (Log-E) and the square root of the eigenvalue (Sqrt-E). As shown in TABLE~\ref{results_on_NU45}, our MG-CAP frameworks are much more accurate than the previous benchmark methods. The Log-E based MG-CAP model achieves more than double the accuracy of the BoVW method (\textit{i.e.,} 88.35\% versus 41.72\% under the training ratio of 10\%, and 90.94\% versus 44.97\% under the training ratio of 20\%). Interestingly, we find that the bilinear pooling performs better than the Log-E based method. We argue that this is because the logarithm of the eigenvalues considerably changes the magnitudes of eigenvalues, especially for small eigenvalues~\cite{sos:li2017second}. This change will reverse the significances of eigenvalues and is harmful to the performance. The Sqrt-E based MG-CAP reasonably avoids this problem and achieves the best classification results.

% \usepackage{multirow}
\begin{table*}[!htbp]
	\centering
	\caption{Comparison of classification accuracy (\%) achieved by our MG-CAP framework, deep learning-based baselines and state-of-the-art methods, where T.R. is the abbreviation of Training Ratio.}
	\begin{tabular}{ccccccc}
		\toprule
		\multirow{2}{*}{}                                            & \multirow{2}{*}{Deep Learning-based Methods}  &  \multicolumn{2}{c}{NWPU-RESISC45}         & \multicolumn{2}{c}{AID}                   & UC-Merced     \\\cline{3-7} 
		&                         & T.R.=10\% & T.R.=20\% & T.R.=20\% & T.R.=50\% & T.R.=80\% \\ \hline
		
		& AlexNet+SVM~\cite{cheng2018deep}             & 81.22$ \pm $0.19          & 85.16$ \pm $0.18          & 84.23$ \pm $0.10          & 93.51$ \pm $0.10          & 94.42$ \pm $0.10          \\
		& GoogLeNet+SVM~\cite{cheng2018deep}           & 82.57$ \pm $0.12          & 86.02$ \pm $0.18          & 87.51$ \pm $0.11          & 95.27$ \pm $0.10          & 96.82$ \pm $0.20          \\
		& VGG-D+SVM~\cite{cheng2018deep}               & 87.15$ \pm $0.45          & 90.36$ \pm $0.18          & 89.33$ \pm $0.23          & 96.04$ \pm $0.13          & 97.14$ \pm $0.10          \\
		
		& MSCP with AlexNet~\cite{he2018remote}         & 81.70$ \pm $0.23          & 85.58$ \pm $0.16          & 88.99$ \pm $0.38          & 92.36$ \pm $0.21          & 97.29$ \pm $0.63          \\
		& MSCP+MRA with AlexNet~\cite{he2018remote}     & 88.31$ \pm $0.23          & 87.05$ \pm $0.23          & 90.65$ \pm $0.19          & 94.11$ \pm $0.15          & 97.32$ \pm $0.52          \\
		
		& MSCP with VGG-D~\cite{he2018remote}         & 85.33$ \pm $0.17          & 88.93$ \pm $0.14          & 91.52$ \pm $0.21          & 94.42$ \pm $0.17          & 98.36$ \pm $0.58          \\
		& MSCP+MRA with VGG-D~\cite{he2018remote}     & 88.07$ \pm $0.18          & 90.81$ \pm $0.13          & 92.21$ \pm $0.17          & 96.56$ \pm $0.18          & 98.40$ \pm $0.34          \\
		& DCNN with AlexNet~\cite{cheng2018deep}     & 85.56$ \pm $0.20          & 87.24$ \pm $0.12          & 85.62$ \pm $0.10          & 94.47$ \pm $0.10          & 96.67+0.10          \\
		& DCNN with GoogLeNet~\cite{cheng2018deep}     & 86.89$ \pm $0.10          & 90.49$ \pm $0.15          & 88.79$ \pm $0.10          & 96.22$ \pm $0.10          & 97.07+0.12          \\
		& DCNN with VGG-D~\cite{cheng2018deep}         & 89.22$ \pm $0.50          & 91.89$ \pm $0.22          & 90.82$ \pm $0.16          & \textbf{96.89}$ \pm $0.10          & 98.93$ \pm $0.10 \\     
		&RTN with VGG-D~\cite{chen2018recurrent}     & 89.90 		   & 92.71		  & 92.44		  & -		
		& 98.96 \\\midrule 
		&\textbf{MG-CAP with Bilinear (ours)}                                                     &{89.42}$\pm${0.19}                     &{91.72}$\pm${0.16}                     &{92.11}$\pm${0.15}                     &{95.14}$\pm${0.12}                       &{98.60}$\pm${0.26} \\
		&\textbf{MG-CAP with Log-E (ours)}                                                     &{88.35}$\pm${0.23}                     &{90.94}$\pm${0.20}                     &{90.17}$\pm${0.19}                     &{94.85}$\pm${0.16}                       &{98.45}$\pm${0.12}                     \\
		&\textbf{MG-CAP with Sqrt-E (ours)}                                                   &\textbf{90.83}$\pm$0.12                   &\textbf{92.95}$\pm$0.13                     &\textbf{93.34}$\pm$0.18                      &96.12$\pm$0.12                     &\textbf{99.0}$\pm$0.10                     \\ \bottomrule  
	\end{tabular}\label{results_overall}
\end{table*}

To adequately evaluate the effectivenesses of our MG-CAP framework, we compare it with state-of-the-art approaches. The results reported in TABLE~\ref{results_on_NU45} and TABLE~\ref{results_overall} are obtained by initialising three granularities and 12 rotations for each granularity. From TABLE~\ref{results_overall}, the VGG-D~\cite{simonyan2014very} based architecture usually achieves a more desirable classification accuracy than GoogLeNet or AlexNet. The Log-E based MG-CAP method obtains 88.35\% accuracy when using 10\% of the training samples, surpassing all of the MSCP based methods~\cite{he2018remote}. When using Bilinear pooling, our MG-CAP achieves an even higher accuracy than the metric learning-based Discriminative CNNs (DCNN)~\cite{cheng2018deep} and is very close to RTN~\cite{chen2018recurrent}. Our Sqrt-E based MG-CAP algorithm obtains 90.64\% and 92.75\% classification accuracy under the different cases, which are the best results so far on NWPU-RESISC45~\cite{datasets:cheng2017Remote}. On the AID~\cite{datasets:xia2016AID}, the Log-E based MG-CAP model also performs very competitively. For example, the classification accuracy exceeds all of the listed architectures with the linear SVM methods~\cite{cheng2018deep}. When using the VGG-D architecture~\cite{simonyan2014very}, the results obtained are slightly lower than DCNN~\cite{cheng2018deep} and significantly lower than MSCP~\cite{he2018remote} and RTN~\cite{chen2018recurrent}. Surprisingly, MSCP with MRA~\cite{he2018remote} achieved more reliable results than DCNN~\cite{cheng2018deep} under the training ratio of 20\%. However, MSCP~\cite{he2018remote} cannot be trained in an end-to-end manner. Again, our MG-CAP with Sqrt-E achieves the best classification accuracy on the AID under the training ratio of 20\%. Specifically, we obtain 93.34\% accuracy, with relative gains of 2.52\% and 1.1\% compared with DCNN~\cite{cheng2018deep} and RTN~\cite{chen2018recurrent}. Although DCNN~\cite{cheng2018deep} performs slightly better than our algorithm under the training ratio of 50\%, our Sqrt-E based algorithm exceeds all linear SVM based methods and obtains competitive performance to MSCP~\cite{he2018remote}. The UC-Merced dataset~\cite{yang2010bag} contains 21 categories, which is comparatively less than other datasets. Using a large number of training samples, all of the listed deep learning approaches achieve similar results. For example, the reported accuracy of the fine-tuned VGG-D~\cite{simonyan2014very} model is 97.14\%, which is on a par with the GoogLeNet based DCNN architecture~\cite{cheng2018deep}. Additionally, our Sqrt-E based MG-CAP achieves 99.0\% classification accuracy, which is the best result among all of the compared methods. 

We show an example of a category-level classification result in Fig.~\ref{mg_cap:nwpu_cm}. Specifically, the accuracies of 43 categories are higher than 80\%. Also, the accuracies of 31 of all 45 categories can exceed 90\%. For the most visually similar categories (\textit{Church} and \textit{Palace}), we achieve substantial improvements of 8\% and 13\% compared with the results reported in~\cite{datasets:cheng2017Remote}.     
\subsection{Ablation Studies}  
\begin{table}[]
	\centering
	\caption{Comparison of accuracy achieved when using different granularities on NWPU-RESISC45~\cite{datasets:cheng2017Remote} with 10\% training ratio.}
	\scalebox{0.9}{
		\begin{tabular}{ccccccc}
			\toprule
			Granularities                                                    &1 &2 &3 &1+2 &2+3 &1+2+3 \\\hline
			\bf{MG-CAP (Log-E)}        &85.50      &86.22      &85.79      &87.13        &86.57        &88.45          \\
			\bf{MG-CAP (Sqrt-E)} &88.63      &89.05      &87.84      &90.17        &89.82        &90.95   
			\\ \bottomrule       
	\end{tabular}}\label{t4}
\end{table}
\subsubsection{Effect of Granularity} 
\begin{figure}[]
	\centering
	\includegraphics[width=0.48\textwidth]{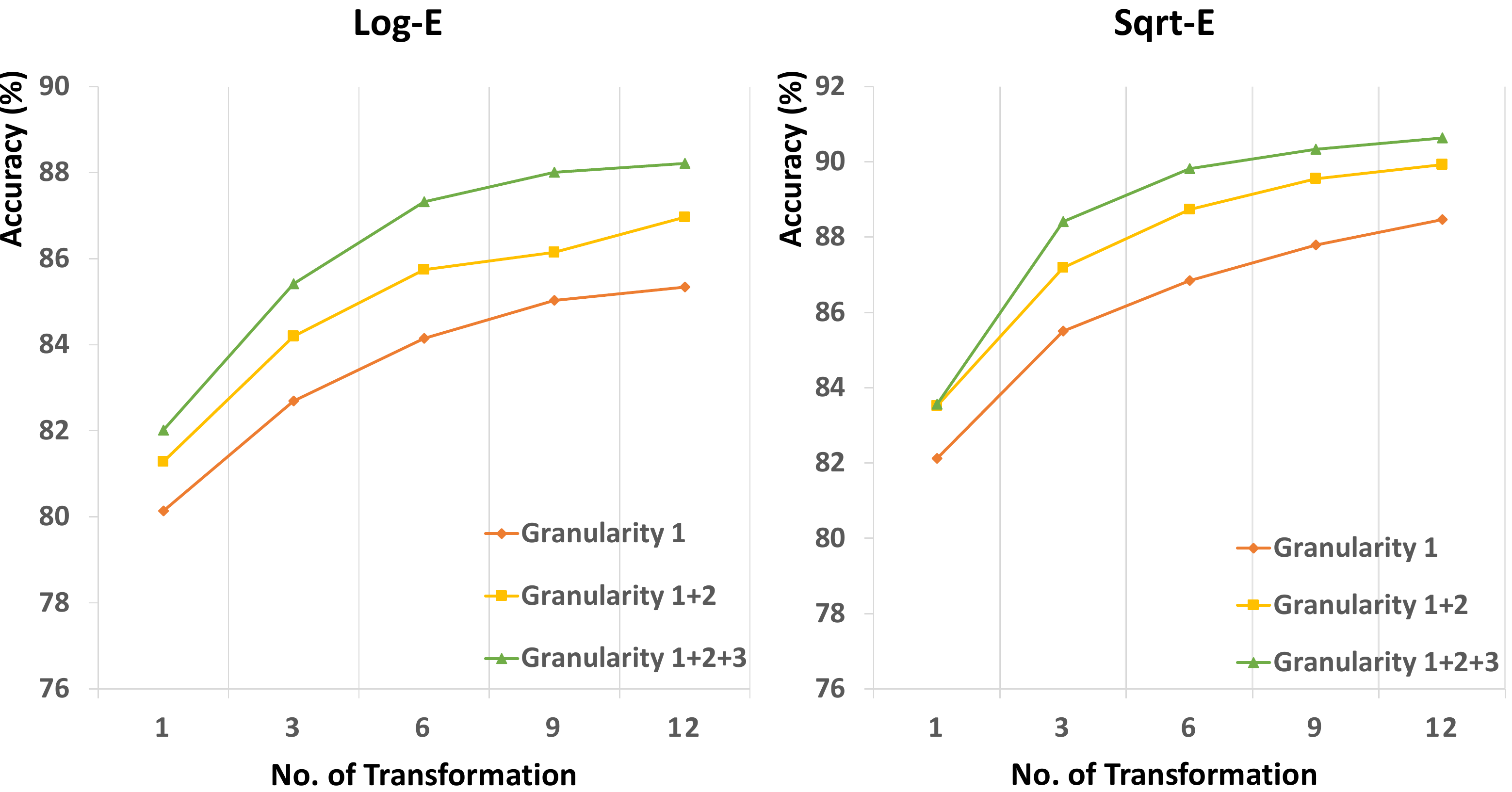}
	\caption{Classification accuracy when using different numbers of transformations.}
	\label{chart}
\end{figure}
\begin{figure*}[]
	\centering
	\includegraphics[width=0.95\textwidth]{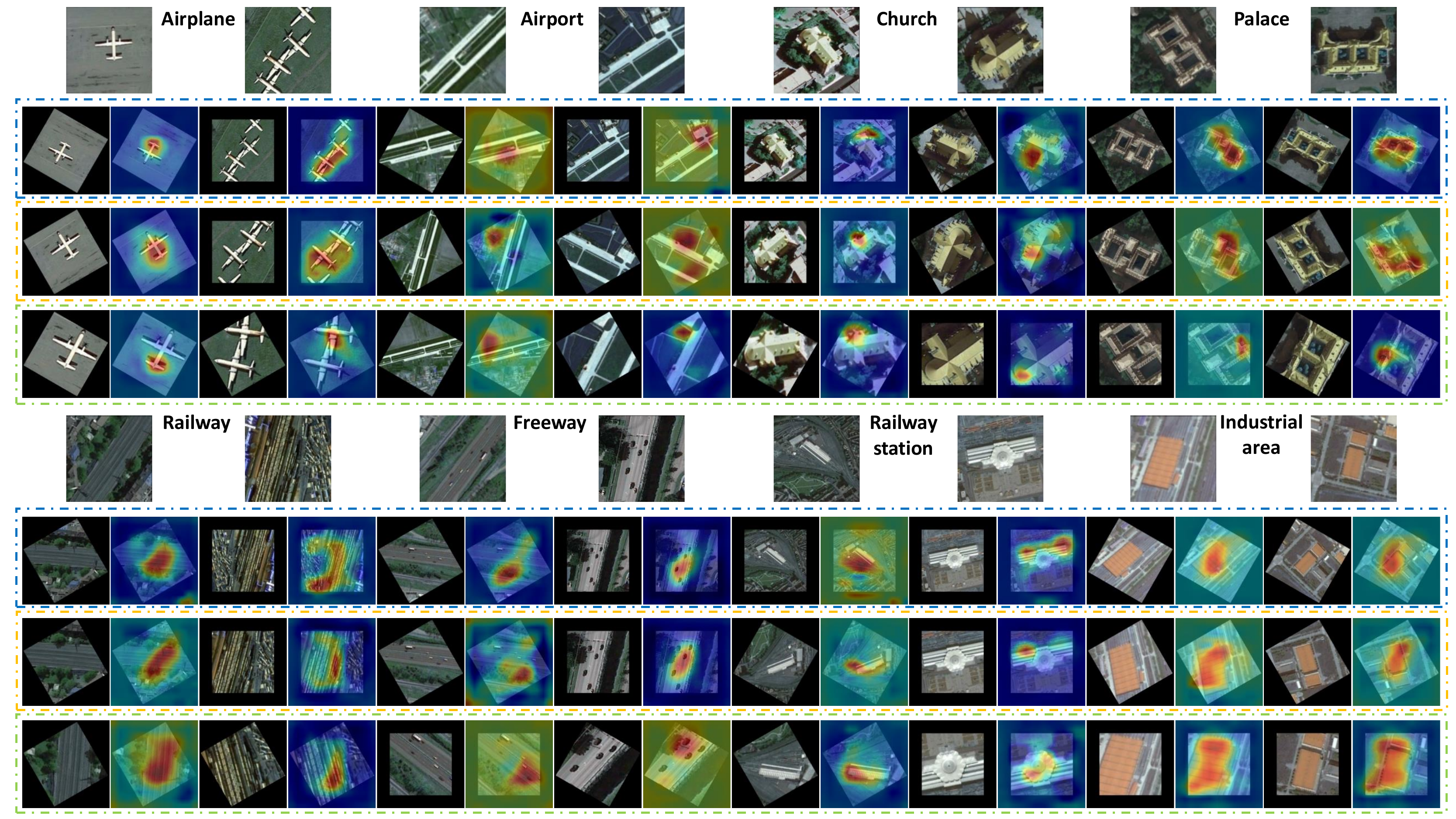}
	\caption{Example visualisation of our MG-CAP model on NWPU-RESISC45~\cite{datasets:cheng2017Remote}, where the blue, yellow and green dashed lines denote the $ 1^{st} $, $ 2^{nd} $ and $ 3^{rd} $ granularity, respectively (best seen in colour).}
	\label{visual}
\end{figure*}   
We construct six different cases to demonstrate the influence of using different granularities. As shown in TABLE~\ref{t4}, the best results are achieved when combining three different granularities. For an individual granularity, the best classification accuracy is achieved by the second granularity while the third granularity is the least accurate. Combining the first and second granularities leads to a better performance than integrating the second and third granularities (\textit{i.e.,} 87.13\% versus 86.57\% with Log-E, 90.17\% versus 89.82\% with Sqrt-E). These results suggest that the finer crop can increase classification accuracy but the overcropping may damage the performance.
\subsubsection{Influence of Transformations}
Fig.~\ref{chart} reflects how the number of transformations affects the final classification accuracy for both the Log-E and Sqrt-E based MG-CAP approaches. As we can see, the classification results increase proportionally with transformations. Notably, the accuracy improves by about 3\% when rotating three times for each granularity. This encourages us to introduce more transformations, but we find that the growth rate becomes relatively stable. As increasing the number of transformations will dramatically raise the memory burden, we set the transformations to 12 for all experiments. 
\subsection{Qualitative Visualisation and Analysis}
In addition to improving accuracy, we are also concerned about the model's interpretability. We provide two approaches to show the canonical appearance and the discriminative part of an image, respectively. The canonical appearance is derived from the Eq.~\ref{eq4} and the corresponding derivative of Eq.~\ref{eq28}. Specifically, the optimal transformation for any grain level can be obtained by:$ \phi=\underset{\phi  \in \Phi} {\mathrm{argmax}} ~f_c((\textbf{G}_s^{\phi})^+) $. To visualise the most discriminative parts of images, we adopt an off-the-shelf algorithm (\textit{i.e.,} Grad-CAM~\cite{selvaraju2017grad}) to display the attention heatmap of the test image. 

We randomly choose several test images from the NWPU-RESISC45 dataset~\cite{datasets:cheng2017Remote} to show the effectiveness of our MG-CAP in terms of learning the canonical appearance and the discriminative features. As shown in Fig.~\ref{visual}, we can see that our MG-CAP tends to orientate visually similar images or image regions in approximately the same direction, and vice versa. For example, canonical appearances learned on different grain levels have almost the same directions for \textit{Church} and \textit{Palace} images. Furthermore, we see that the canonical appearance has changed for the third granularity of the \textit{Palace} images. This is because of the main object is only partially present in this granularity. Besides, we notice that the canonical transformation barely changes in some images, such as \textit{Railway}, \textit{Freeway} and \textit{Railway station}. This is because the texture is visually similar for different granularities. In contrast, the corresponding attention heatmap presents diversity due to the changes in image content. 
\begin{table}[]
	\centering
	\caption{Comparison of complexity and inference time, where \textit{n} denotes the number of streams. Both sides of a backslash indicate the inference time on a CPU and GPU.}
	\begin{tabular}{cccc}
		\toprule
		& Model Complexity  & \begin{tabular}[c]{@{}c@{}}\#Params\\ (MB)\end{tabular} & \begin{tabular}[c]{@{}c@{}}Inference Time \\ (sec/img)\end{tabular}\\\midrule
		RTN~\cite{chen2018recurrent}      &\textit{ O(n(LocNet+BilinearVGG))} &68.69                                               &0.434/0.073                 \\
		\begin{tabular}[c]{@{}c@{}}\textbf{MG-CAP}\\ (Norm-E)\end{tabular}  & \textit{O(n(CovarianceVGG))}         &55.99                                         &1.837/0.216                   \\ \bottomrule 
	\end{tabular}\label{mg_cap:complexity}
\end{table}

We also compare the time complexity of our algorithm with the multi-stream based RTN method~\cite{chen2018recurrent}. We replicate the RTN~\cite{chen2018recurrent} model using the released source code. Then, we conduct experiments using a PC with a 6-core Intel\textsuperscript{\textregistered} Core\textsuperscript{TM} i7-9800X@3.80 GHz CPU and a GeForce RTX 2080Ti GPU. From TABLE~\ref{mg_cap:complexity}, we can see that the model complexity of RTN~\cite{chen2018recurrent} is \textit{O(n(LocNet+BilinearVGG))}, which is more complicated than our MG-CAP model. Specifically, RTN~\cite{chen2018recurrent} needs localisation networks to be recursively applied in order to predict the transformation parameters. In terms of model parameters, our MG-CAP based method requires 55.99 MB of memory, while RTN~\cite{chen2018recurrent} takes up an extra 12.7 MB. Although RTN~\cite{chen2018recurrent} has a shorter inference time, our GPU-based MG-CAP only needs 0.216 seconds for the predictions. In addition, our EIG-decomposition function based MG-CAP algorithm can significantly improve the classification accuracy. Besides, the inference time of our MG-CAP model has been significantly reduced since we implemented the stable GPU version of the EIG-decomposition function. The cost of the matrix decomposition function can be further decreased by learning compact representations of Gaussian covariance matrices or using powerful GPUs.  
\section{Conclusion}    
In this paper, we proposed a novel MG-CAP framework to solve the large visual-semantic discrepancy and variance problems in RSSC. Our framework was built on a multi-granularity fashion and can automatically learn the latent ontologies in remote sensing scene datasets. At each specific grain level, a maxout-based Siamese style architecture was employed to discover the canonical appearance and extract the corresponding CNN features. We collected second-order statistics of standard CNN features and then transformed them into Gaussian covariance matrices. We adopted suitable matrix normalisations to improve the discriminative power of second-order features. More importantly, we offered solutions that allow the EIG-decomposition function to be well supported by GPU-acceleration. The whole framework can be trained in an end-to-end manner. In the future, we will investigate ways to compress the model parameters while maintaining the classification performance. 
\bibliographystyle{IEEEtran}
\bibliography{egbib}
\end{document}